\documentclass[fleqn,10pt]{wlscirep}
\usepackage[utf8]{inputenc}
\usepackage{comment}
\usepackage[T1]{fontenc}

\usepackage[pagewise]{lineno}  

\usepackage{tikz}
\usetikzlibrary{positioning}
\usetikzlibrary{fit}
\usetikzlibrary{calc}
\usetikzlibrary{decorations.pathreplacing,angles,quotes}

\title{RainShift: A Benchmark for Precipitation Downscaling Across Geographies}

\author[1,2,+]{Paula Harder}
\author[3,*,+]{Luca Schmidt}
\author[1]{Francis Pelletier}
\author[3]{Nicole Ludwig}
\author[2]{Matthew Chantry}
\author[2]{Christian Lessig}
\author[4]{Alex Hernandez-Garcia}
\author[1,4]{David Rolnick}

\affil[1]{Mila Quebec AI Institute, Montreal, Canada}
\affil[2]{European Centre for Medium Range Weather Forecasts (ECMWF), Bonn, Germany}
\affil[3]{Cluster of Excellence Machine Learning, University of Tübingen, Tübingen, Germany}
\affil[4]{McGill University, Montreal, Canada}
\affil[*]{luca-marie.schmidt@mila.quebec}
\affil[+]{these authors contributed equally to this work}

\begin{abstract}
Earth System Models (ESM) are our main tool for projecting the impacts of climate change. However, running these models at sufficient resolution for local-scale risk-assessments is not computationally feasible. Deep learning-based super-resolution models offer a promising solution to downscale ESM outputs to higher resolutions by learning from data. Yet, due to regional variations in climatic processes, these models typically require retraining for each geographical area—demanding high-resolution observational data, which is unevenly available across the globe. This highlights the need to assess how well these models generalize across geographic regions. To address this, we introduce RainShift, a dataset and benchmark for evaluating downscaling under geographic distribution shifts. We evaluate state-of-the-art downscaling approaches including GANs and diffusion models in generalizing across data gaps between the Global North and Global South. Our findings reveal substantial performance drops in out-of-distribution regions, depending on model and geographic area. While expanding the training domain generally improves generalization, it is insufficient to overcome shifts between geographically distinct regions. We show that addressing these shifts through, for example, data alignment can improve spatial generalization. Our
work advances the global applicability of downscaling methods and represents a step toward reducing inequities in access to high-resolution climate information.
\end{abstract}

\begin{document}

\flushbottom
\maketitle

\thispagestyle{empty}

\section*{Introduction}
High-resolution climate projections are crucial for planning effective responses to extreme weather and climate events. With ongoing climate change, extreme events such as floods, droughts, and heatwaves are expected to become more frequent and severe, threatening infrastructure, agriculture, energy systems, and public health \cite{seneviratne2021weather}. Among these, precipitation extremes are particularly destructive, causing floods, landslides, and soil erosion; and are projected to intensify at local scales \cite{ombadi2023warming, xiong2024climate}. Yet, accurately modeling precipitation extremes remains difficult due to their high spatial and temporal variability and the non-linear, multi-scale processes involved \cite{maraun2010precipitation, pendergrass2017precipitation}.

Earth System Models (ESMs) are the primary tools for understanding the Earth’s climate system and projecting future conditions. By representing physical, chemical, and biological processes, they simulate interactions between climate, the carbon cycle, ecosystems, and human activities. However, the spatial resolution of ESMs—typically around 100 km—is too coarse to resolve small-scale processes. Instead, sub-grid processes are approximated through parameterizations that estimate their average influence. A critical example is deep convection, a major driver of precipitation and a major cause of extreme rainfall, such as flash floods and landslides \cite{fosser2024convection}. As these processes occur at spatial scales finer than those resolved by ESMs, the models are unable to accurately represent these events \cite{maraun2010precipitation}.

To address this limitation, downscaling methods are used to increase the spatial resolution of ESM output. There are two families of approaches for downscaling: dynamical downscaling and statistical downscaling.
Dynamical downscaling uses a high-resolution regional climate model driven by ESM boundary conditions to simulate fine-scale processes within a limited area of interest.
In contrast, statistical downscaling techniques learn empirical relationships between large-scale predictors and local-scale observations from training data and apply these learned relationships to predict localized climate outputs. Compared to dynamical downscaling methods, statistical downscaling methods are more computationally efficient but require high-resolution data for training.

Recently, deep learning methods have shown promise for statistical downscaling, leveraging advances in computer vision—particularly super-resolution techniques. Early approaches used convolutional neural networks (CNNs) to learn deterministic mappings from coarse to high resolution \cite{vandal2017}, with architectures such as super-resolution CNNs, U-Nets \cite{Sha2020DeepLearningBasedGD, https://doi.org/10.1002/met.1961}, and ResNets \cite{10.1145/3394486.3403366, 8588749} being widely applied to climate downscaling.
More recent work has shifted toward probabilistic frameworks, leveraging generative models to represent the inherent uncertainty in the downscaling task. Generative adversarial networks (GANs) \cite{Goodfellow2014GenerativeAN}, especially conditional generative adversarial networks (cGANs) \cite{Goodfellow2014GenerativeAN} and stabilized variants like Wasserstein GANs \cite{Harris2022AGD, cooper2023analysiscgangenerativedeep} are popular choices for downscaling. More recently, diffusion models have also shown strong performance in modeling complex, high-dimensional distributions \cite{SohlDickstein2015DeepUL, Mardani2023GenerativeRD, Wan2023DebiasCS, addison2024machinelearningemulationprecipitation, ling_diffusion_2024}.
Among all climate variables, precipitation is particularly challenging to model and forecast due to its stochastic, high-frequency, spatial and temporal variability. Capturing these fine-scale variations and inherent uncertainties would make generative approaches successful for precipitation downscaling. 

Unlike dynamical downscaling, statistical downscaling algorithms are not restricted to specific data sources or geographic regions. 
By construction, statistical downscaling relies on statistical relationships learned during the training phase, allowing it to be applied to new regions at inference time. However, this transferability hinges on the assumption that both predictor and predictand distributions, as well as their statistical relationships, remain stationary across space.
In practice, this assumption does not hold due to substantial geographic variability in topography, climatic conditions and processes. For example, precipitation in equatorial regions is strongly driven by convection and, therefore, substantively different than in Europe and North America. Consequently, statistical downscaling models often require retraining for each target region, which relies on the availability of high-resolution observational data.

\begin{figure}[ht]
\centering
\includegraphics[width=12cm]{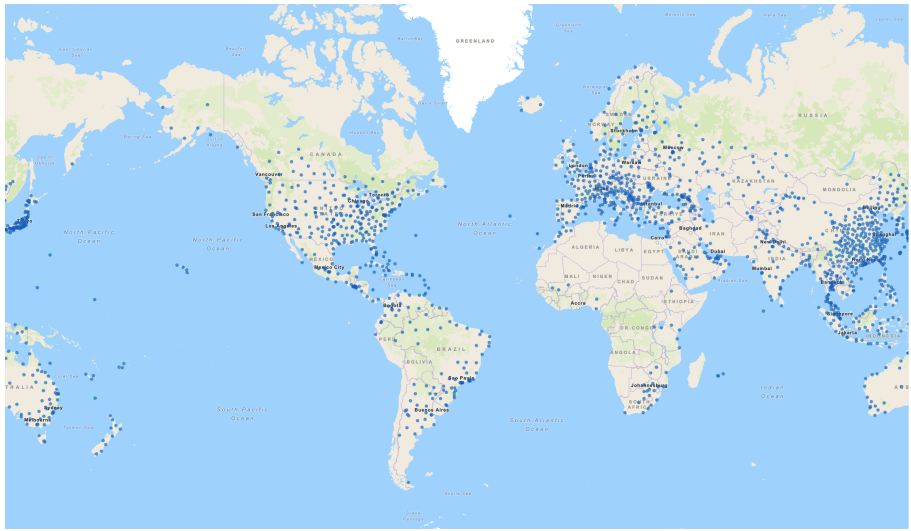}
\caption{\textbf{Map of ground-based radar stations.} The map shows the availability of precipitation data, with each blue dot representing a station. Coverage is relatively high in the Global North and comparatively low across the Global South. Image from the Tropical Globe radar database \cite{tropical_globe_radar_database}.}
\label{fig:radar}
\end{figure}

Despite the large amount of weather and climate datasets, high-quality observations are unevenly distributed globally. Ground-based radar and gauge data, essential for training and validating downscaling models, are particularly sparse in many parts of the Global South (see Figure \ref{fig:radar}). Yet, it is also these regions that are often the most exposed and vulnerable to climate change and extreme weather events like heavy rainfall and flooding \cite{xiong2024climate}. This global imbalance in data availability—coupled with heterogeneous regional climate processes—presents challenges in generalization of deep learning-based downscaling models.

To address these challenges, we introduce RainShift,  a large-scale global benchmark and dataset designed to evaluate the geographical generalization of deep learning-based downscaling. RainShift defines test scenarios where models are trained on subsets of data-rich regions and tested on regions with scarce availability of high-resolution observations. 
The dataset is built from ERA5 reanalysis and IMERG satellite precipitation data. 
We establish baseline results by evaluating state-of-the-art models for probabilistic precipitation downscaling, including GANs and diffusion-based architectures. RainShift is intended to support the development of approaches that generalize to low-data regions, particularly in underrepresented areas such as the Global South. Our main contributions are the following:

\begin{itemize}
    \item We frame the task of downscaling across a geographically varying data distribution, based on a critical gap in Earth system modeling. 
    \item We introduce the RainShift benchmark dataset, along with tools for expanding it to new regions, data loaders, training pipelines and evaluation frameworks, within an accessible library to facilitate further research. 
    \item We evaluate a variety of state-of-the-art machine learning downscaling models on RainShift and find substantial variation in spatial generalization across models and regions; we show that data alignment techniques can improve performance in cases where generalization is limited by strong geographic distribution shifts.
\end{itemize}

\section*{Results}
\begin{figure} \label{fig: setup}
\begin{center}
\input{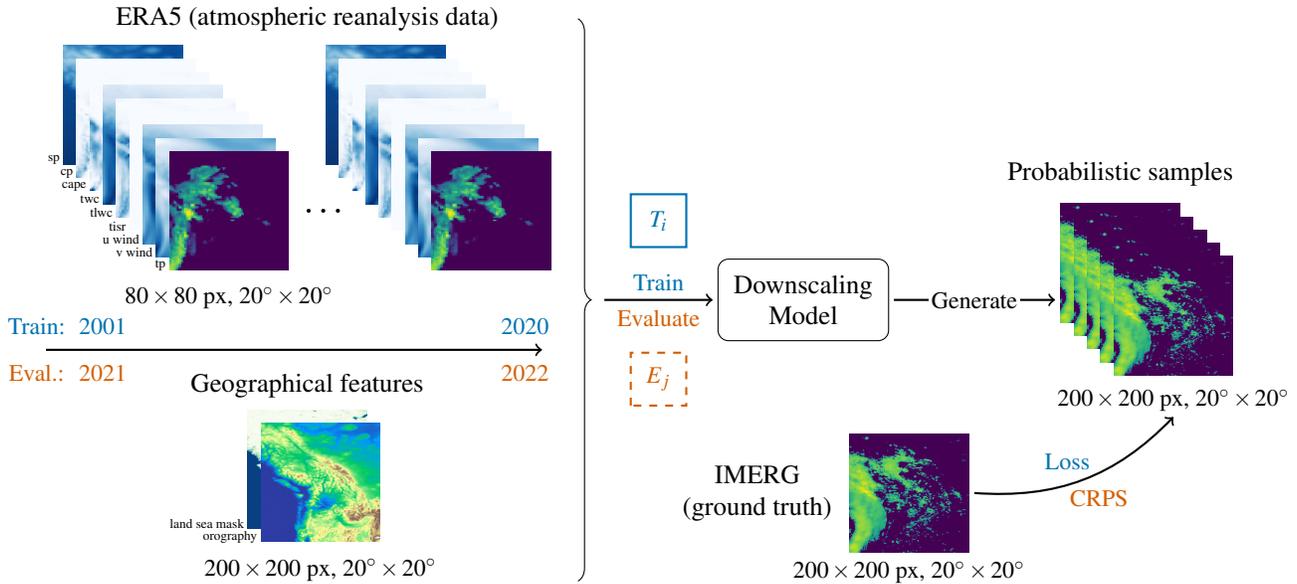}
\end{center}
\caption{\textbf{Graphical summary of RainShift setup.} The inputs of the downscaling model are a combination of ERA5 time series data and geographical features. The downscaling model is then able to generate probabilistic samples. For training, we sample from geographic areas $T_1,\ldots,T_{12}$ and years 2001--2020, and compare the generated samples with the ground truth target IMERG to compute the loss. For evaluation, we use areas $E_1,\ldots,E_6$, and years 2021--2022.}
\label{fig:setup}
\end{figure}

\begin{figure}[ht]
\centering
\includegraphics[width=10cm]{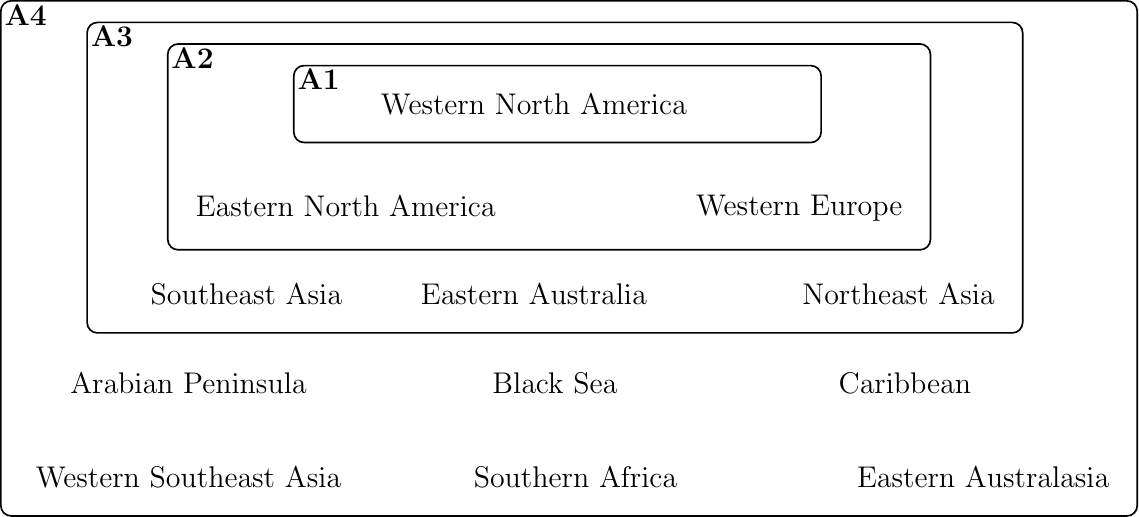}
\caption{\textbf{Illustration of training configurations.} The training configurations $A1,...,A4$ are composed of progressively larger subsets of the 12 selected training regions located in the Global North. The choice of regions is guided by availability of high-resolution observational data and inspired by existing works \cite{cooper2023analysiscgangenerativedeep, prasad2024evaluating}.} 
 \label{fig:training_cfgs}
\end{figure}

\begin{figure}[ht]
\includegraphics[width=\textwidth]{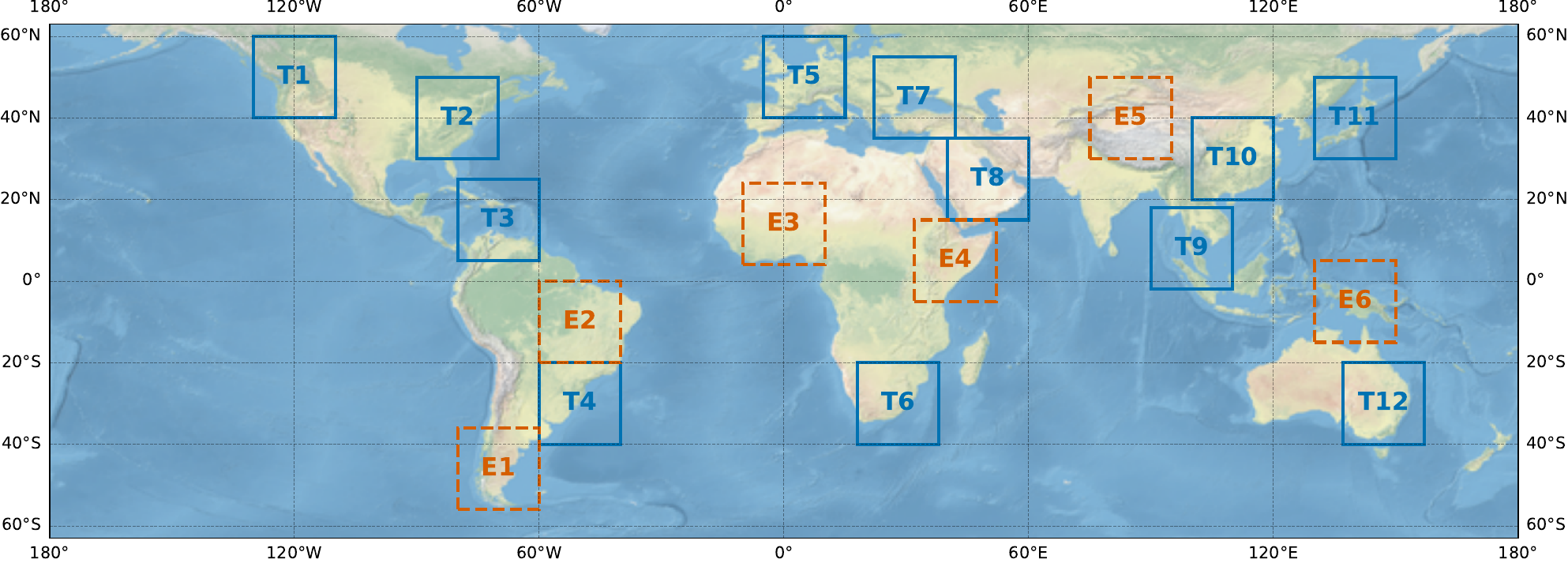}
\caption{\textbf{Illustration of location splits and training configurations.} Patches $T_{1},...,T_{12}$ represent training regions and patches $E_{1},...,E_{6}$ correspond to evaluation areas that are used within $4$ sub-tasks, simulating different scenarios that correspond to varying levels of data availability.}
\label{fig:location_split}
\end{figure}

We train a variety of downscaling models in scenarios representing data-abundant regions and then evaluate their performance in data-sparse regions, mainly in the Global South. The downscaling task consists of learning a mapping from coarser resolution reanalysis data (ERA5) to higher resolution satellite-based  precipitation (IMERG), as illustrated in Figure \ref{fig:setup}. These globally consistent data sources enable a controlled benchmark setup. To evaluate spatial generalization, we define 12 training regions and 6 evaluation regions. We combine the training regions into four progressively larger training scenarios ($A1,...,A4$) to simulate varying levels of observational coverage (see Figure \ref{fig:training_cfgs}); their geographic distribution, along with that of the evaluation regions, is shown in Figure~\ref{fig:location_split}. 

The analysis includes several state-of-the-art downscaling approaches: ResNets \cite{Harris2022AGD} (a deterministic approach), Wasserstein GAN with gradient penalty \cite{Arjovsky2017WassersteinG} (a probabilistic approach), and a diffusion-based method \cite{watt2024generative} (also probabilistic).
We also include a simple bilinear interpolation as a baseline. 
We evaluate model performance with respect to spatial generalization using the continuous ranked probability score (CRPS), considering out-of-distribution performance both in absolute terms and relative to in-distribution evaluation.
Our analysis investigates key factors affecting generalization, including the downscaling model, the geographic region, and size of the training domain. To address the observed performance drops in unseen regions, we propose a simple data alignment approach based on quantile mapping to improve the spatial generalization ability of the evaluated models.

\subsection*{Added value of downscaling models}  
All learned models demonstrate some generalization to unseen target regions. As shown in Figure \ref{fig:rel_bilinear}, they achieve consistent improvements (except for two outlier values) on the ERA5 interpolation, which serves as a baseline input compared to IMERG. This trend holds across geographical areas. The consistency of these improvements across diverse target areas shows the general validity and robustness of the downscaling approach, indicating that such models can be used to downscale coarse-resolution inputs even when applied to regions not seen during training. However, the magnitude of relative improvement varies greatly between target regions, with relative CRPS reductions ranging between approximately 30\% and 50\%.

\subsection*{Probabilistic models outperform deterministic ones}  
Across almost all evaluation regions, probabilistic generative approaches, both GAN and diffusion models, consistently outperform the deterministic ResNet model. In some cases, ResNet even shows unstable behavior, such as in region E5 (Tibetan Plateau) under configuration A1, where the CRPS value becomes unreliable due to numerical instabilities that occur exclusively during inference. The issue appears to be the result from a combination of limited training data and a pronounced distribution shift between training and target areas. Specifically, the Tibetan Plateau has much lower precipitation compared to the A1 (Western North America) training region. In preliminary experiments in Table \ref{tab:qm_results}, we show that aligning the data distributions between training and target areas effectively resolves this issue.

Both GAN and diffusion show stable and consistent improvements over bilinear interpolation across regions and training configurations. In the largest training setup (A4), GAN and diffusion achieve comparable absolute (see Figure \ref{tab:crps_results}) and relative improvements (see Figure \ref{fig:rel_bilinear}) over the baseline. However, the two models show differences in how they respond to extending the training domain. The diffusion model already generalizes well when trained on a smaller training domain (A1), whereas the GAN shows more pronounced improvements with increases in training data. This suggests that diffusion models are more robust to limited training data, while GANs benefit more from larger, more diverse training datasets.

\subsection*{Expanding the training area is not always beneficial}  
Overall, model performance generally improves when expanding the training areas from A1 to A4 (see Figure \ref{fig:rel_bilinear} and Table \ref{tab:crps_results}). 
In some regions---such as Cape Horn, Amazon Basin, West Africa and Melanesia---expanding the training domain leads to clear improvements in CRPS scores. However, this trend is less consistent in other regions. In the Horn of Africa and the Tibetan Plateau adding more training data does not necessarily lead to better generalization ability. Furthermore, the benefits of expanding training area tend to diminish with larger training domains. This suggests that increasing training data alone does not guarantee continued improvements in spatial generalization.

\subsection*{In-distribution training does not always lead to the best performance} 
For most regions and models, in-distribution training leads to the best CRPS scores (see Table \ref{tab:crps_results}). Consequently, we observe substantial relative performance drops in Figure \ref{fig:rel_target} when models are trained out-of-distribution. For the smallest training setup (A1), these drops can reach up to 30\%, and even with the largest setup (A4), declines of up to 17\% remain.
The diffusion model generally performs best when trained in-distribution. Although larger training domains improve its performance, they typically do not match the performance of on-target training. For the GAN model, performance varies more strongly across regions. In the Amazon Basin and Cape Horn, the size of the training domain appears more important than its overlap with the target region, with A3 and A4 performing as well or even better than in-distribution training. In the remaining four regions, however, models trained directly on the target region still perform best. The ResNet model shows more mixed results, with clear gains from on-target training in some regions, but less consistent patterns in others. Overall, these findings show that excluding the target region from training can significantly degrade model performance, depending on the region, model type, and training domain size. This highlights the importance of developing techniques that improve generalization to unseen regions.

\subsection*{Geographical variation challenges generalization}  
Bilinear interpolation of input precipitation reveals substantial differences in prediction accuracy between target regions. The interpolation error of ERA5, measured by CRPS/MAE (see Table \ref{tab:crps_results}), is particularly high in the regions E2 (Amazon Basin) and E6 (Melanesia), which also have the highest average precipitation among all regions (see Table \ref{tab:stats}). Across regions, we observe a strong positive correlation between mean precipitation and interpolation error: areas with higher precipitation tend to have larger prediction errors. 

A similar trend is reflected in the performance of deep learning models, where regions with higher precipitation remain more difficult to predict, both in absolute (see Table \ref{tab:crps_results}) and in relative improvement (see Figure \ref{tab:crps_results}) over the interpolation baseline. This finding is consistent across different model architectures (GANs, diffusion model, and ResNet) and across different training scenarios (A1-A4). These results suggest that regions with higher precipitation, likely associated with greater precipitation variability, pose greater challenges for spatial generalization in downscaling models.

\subsection*{Geographical factors dominate generalization performance}
While differences between model architectures are evident—generative models such as GANs and diffusion models consistently outperform ResNet---there is relatively little difference between the performance of GAN and diffusion models themselves. However, the dominant factor influencing generalization performance is the geographical area considered. Across all models, distributional shifts across regions with large climatic differences, such as between the Amazon Basin and Melanesia, are consistently more challenging to predict. This indicates that spatial generalization is more constrained by geographic and climatic variability than by architectural choices alone, highlighting the importance of addressing such shifts through improved domain alignment or region-aware modeling strategies. In the preliminary results shown in Table \ref{tab:qm_results}, we see that simple distributional correction techniques such as quantile mapping used to align the feature distribution of training and target regions can greatly improve performance. This is achieved by matching the cumulative distribution functions of the precipitation inputs in the target region to that of the training region. Figure \ref{cdf} illustrates the distributional discrepancies, which are particularly pronounced in regions like Tibetan Plateau and Melanesia. After applying quantile correction, the CDFs are much more closely aligned, leading to improved performance across nearly all regions (see Table \ref{tab:qm_results}).

\begin{figure}[ht]
\centering
\includegraphics[width=\textwidth]{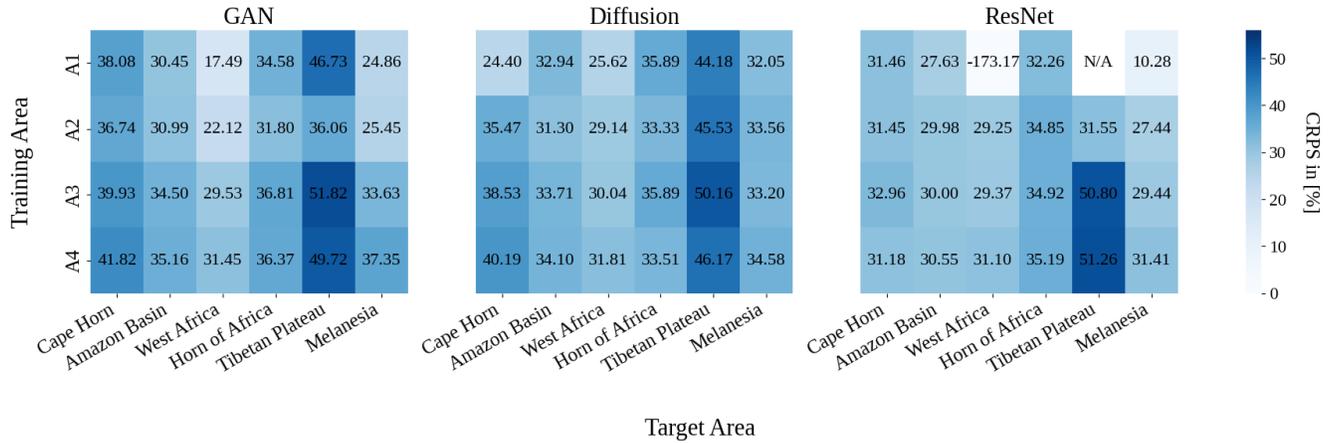}

\caption{\textbf{Heatmap of \% improvement relative to interpolation.} Change in CRPS (lower better) in $[\%]$ for each model relative to bilinear interpolation. $A_{1},...,A_{4}$ represent hierarchical training scenarios with progressively more high-resolution data, from training on a single region (A1) to using several training regions across the Global North (A4). Positive values show improvements over the interpolation baseline. Model performance generally improves when expanding the training areas from A1 to A4, but this trend depends strongly on the geographical region. 
The N/A value indicates an instance where the CRPS value is not reliable due to numerical instabilities during inference, likely driven by large distributional differences between training and target regions.}  
\label{fig:rel_bilinear}
\end{figure}

\begin{figure}[ht]
\centering
\includegraphics[width=\textwidth]{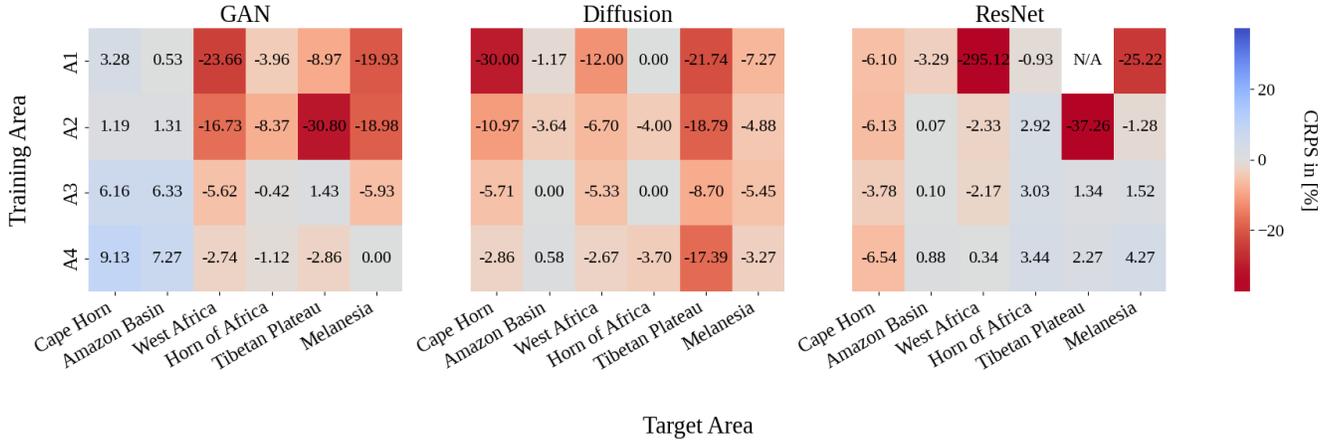}
\caption{\textbf{Heatmap of \% performance drop between in and out-of-distribution training.} Change in CRPS (lower better) in $[\%]$ for each model relative to training the model directly on the target regions. $A_{1},...,A_{4}$  represent hierarchical training scenarios with progressively more high-resolution data, from training on a single region (A1) to using several training regions across the Global North (A4). Many regions show large negative values, indicating substantial drop in performance when evaluating on regions out-of-distribution, highlighting challenges in spatial generalization. 
The N/A value indicates an instance where the CRPS value is not reliable due to numerical instabilities during inference, likely driven by large distributional differences between training and target regions.}  
\label{fig:rel_target}
\end{figure}

\begin{table*}[ht] 
\vskip 0.15in
\begin{center}
\begin{small}
\begin{tabular}{ l l c c c c c c}
\toprule
 & & \multicolumn{6}{c}{Target Areas} \\
Model & QM & E1 & E2 & E3 & E4 & E5 & E6 \\ & & Cape Horn & Amazon Basin & West Africa & Horn of Africa & Tibetan Pl. & Melanesia \\
\toprule
DM & No  &  \textbf{0.091} & \textbf{0.182} & {0.084} & \textbf{0.054} & {0.028} & \textbf{0.295}\\
DM & Yes &   0.092   &    \textbf{0.182}   &   \textbf{0.077}   &   \textbf{0.054}    &  \textbf{0.024}     &    {0.310}   \\
\midrule
GAN       & No  & \textbf{0.075} & \textbf{0.179} &0.093 & \textbf{0.055} &0.027& 0.326  \\
GAN       & Yes &  0.093    &   \textbf{0.179}  &    \textbf{0.080}   &   \textbf{0.055}    &   \textbf{0.024}     &  \textbf{0.310}     \\
\midrule
ResNet & No  & \textbf{0.083}& 0.187 &0.308 & 0.057 & 18113.558& 0.389 \\
ResNet & Yes &  0.093 & \textbf{0.181}& \textbf{0.079}& \textbf{0.055}& \textbf{0.025}& \textbf{0.310} \\
\bottomrule
\end{tabular}
\end{small}
\end{center}
\caption{\textbf{Impact of quantile-based domain alignment on model accuracy.} CRPS (lower values are better) of models trained on scenario A1 (Western North America) and evaluated on the designated target regions for precipitation in [mm/h] over test years $2021 - 2022$. Results show model accuracy with and without applying quantile mapping (QM) to align the input distributions between training and evaluation regions. All predictions are transformed back to the original target domain data range and re-normalized before computing the metrics. Applying quantile mapping equals or improves performance across most models and regions (except Cape Horn) demonstrating the potential of data alignment techniques to enhance spatial generalization under geographical distribution shifts.}
\label{tab:qm_results}
\vskip -0.1in
\end{table*}

\section*{Discussion}
This paper introduces the RainShift benchmark to evaluate the out-of-distribution generalization of deep learning models for precipitation downscaling across geographically distinct regions. Our dataset is constructed from paired reanalysis inputs and satellite-based precipitation targets. It covers twelve training areas with dense observational coverage and six evaluation areas located in the Global South, where high-quality observations are sparse. The benchmark task evaluates a model’s ability to learn from high-resolution training data and generalize to unseen regions, a crucial requirement to the real-world deployment of these models.

Our results show that all evaluated models add substantial value over raw input precipitation, with probabilistic generative approaches (GAN and diffusion) showing better performance than deterministic ones, both within training domains and in generalizing to unseen geographies. However, performance still degrades (up to 37\%) when applied to unseen regions, highlighting current limitations in cross-regional generalization. Our results indicate that spatial generalization is more strongly limited by geographic variability and distribution shifts than by differences in model architecture or training data size, indicating that addressing these shifts is a critical path forward.

To advance progress in this area, we frame cross-regional generalization as a central challenge for current downscaling models and encourage methodological innovation to address it. Our results show that simple data alignment techniques, such as quantile-based input alignment, can enhance model generalization across regions. Beyond this, other promising directions in this regard could include, for example, the use of location-aware embeddings as auxiliary inputs \cite{Klemmer2023SatCLIPGG}, or unsupervised domain adaptation methods that align feature distributions between labeled and unlabeled target domains \cite{Wang31122022}.
Further strategies such as meta-learning \cite{rußwurm2020metalearning}, which enables faster adaptation to new regions with limited data, and the integration of physical constraints to improve model robustness \cite{harder2023hard}, also have potential to improve spatial generalization.

Our choice of training and evaluation regions is guided by the availability of high-resolution observations. To facilitate development and deployment of new methods, we use a globally homogeneous dataset, IMERG, as the target. However, a promising avenue for future work is to expand the benchmark to incorporate additional, more localized sources, such as precipitation radar data from individual countries. Models trained on such data could then be applied to regions without high-quality observational data available, including the Global South. 
More broadly, the RainShift benchmark is designed to provide a framework for evaluating how well models generalize to regions most in need of accurate, high-resolution climate information---bridging the gap between highly localized model development and global applicability.

\section*{Methods}
We introduce the RainShift dataset, along with its preprocessing pipeline, baseline models, and evaluation framework. All components are fully reproducible and will be made publicly available upon acceptance.

\subsection*{RainShift dataset}\label{data}
The RainShift dataset builds on three global data sources: atomospheric reanalysis (ERA5), satellite-based precipitation estimates (IMERG), and two invariant geographical features---land-sea mask and orography. The use of globally consistent satellite and reanalysis data enables a controlled benchmark setup that is essential for evaluating how well downscaling models generalize spatially. Using satellite-derived precipitation as the target allows for evaluation in data-sparse regions like the Global South, while limiting confounding differences due to differences between data products.

\subsubsection*{Input data}
As low-resolution input data, we use ERA5 \cite{era5}, the fifth-generation atmospheric reanalysis product of the European Center for Medium-Range Weather Forecasts (ECMWF). Reanalysis data are the result of combining historical observations with Earth system models through data assimilation to obtain global estimates of the observed climate.
It provides hourly global data at $0.25^\circ \times 0.25^\circ$ resolution (approximately $25~\mathrm{km}$ per pixel in mid-latitudes) on a regular latitude-longitude grid and spans years from 1950 to the present. For compatibility with IMERG, we use data from 2001 onward. 

We select nine input variables as shown in Table \ref{tab:variables}, based on meteorological relevance to predict subgrid rainfall variability guided by the ecPoint model \cite{Hewson2020ALP} and domain-specific knowledge discussed elsewhere \cite{Harris2022AGD}. 

\begin{table}[ht]
\centering
\begin{small}
\begin{tabular}{ll}
\toprule
\textbf{Variable} & \textbf{Description} \\
\midrule
tp & Total precipitation \\
cp & Convective precipitation \\
cape & Convective potential energy \\
twc & Total water content \\
tlwc & Total liquid water content \\
sp & Surface pressure \\
tisr & Top-of-the-atmosphere incident solar radiation \\
u & Eastward wind velocity at 700 hPa \\
v & Northward wind velocity at 700 hPa \\
\bottomrule
\end{tabular}
\end{small}
\caption{\textbf{List of atmospheric variables used as predictors.}}
\label{tab:variables}
\end{table}

We additionally include geographic covariates at $0.1^\circ$ resolution: (i) A land-sea mask indicating the land fraction per pixel, and (ii) an elevation map (geopotential height at surface).

\subsubsection*{Target data}
The Integrated Multi-satellite Retrievals for GPM (IMERG) \cite{huffman_imerg_2014} is a product of NASA’s Global Precipitation Measurement (GPM) mission and serves as high-resolution target data. IMERG provides precipitation estimates based on the GPM satellite constellation and additional observations such as gauge data. IMERG has full coverage between $60^\circ \text{N} $ and $ 60^\circ \text{S}$ at $0.1^\circ$ resolution (about $10~\mathrm{km}$ per pixel) on a regular latitude-longitude grid. We use the IMERG V07 Final Run product \cite{huffman_imerg_v07}, averaging its half-hourly data to hourly to match ERA5’s temporal resolution. Data was accessed via NASA Goddard’s GES DISC.

\subsubsection*{Summary statistics}
Table \ref{tab:stats} shows the average amount of precipitation in mm/h for each target region in low- and high-resolution datasets. We observe a strong correlation between the error and mean precipitation of a region, i.e. regions with higher precipitation tend to have higher prediction errors. 

\begin{table}[htb] 
\vskip 0.15in
\begin{center}
\begin{small}
\begin{tabular}{lcccccccc} 
\toprule
& & &\multicolumn{6}{c}{Target Areas}\\
 &  & & E1 & E2 & E3 & E4 & E5 & E6 \\
  & &  & Cape  & Amazon & West  & Horn & Tibetan  & Melanesia\\
  & & &  Horn & Basin & Africa & of Africa &  Plateau & \\
  \toprule
 Mean precip. & ERA5 &  & 0.11 & 0.18 & 0.07 & 0.06 & 0.04 & 0.33\\
 Mean precip. & IMERG & & 0.10 & 0.18 & 0.08 & 0.05 & 0.02 & 0.31  \\
\bottomrule
\end{tabular}
\end{small}
\end{center}
\caption{\label{tab:stats}
\textbf{Mean precipitation values.} Values are in mm/h averaged over each evaluation region using years 2021-2022.}
\vskip -0.1in
\end{table}

\subsubsection*{Data processing} \label{sec:data-proc}
The datasets are downloaded and preprocessed for consistency. Global data is divided into subregions and stored in the ML-friendly Zarr format \cite{alistair_miles_2020_3773450}. Zarr archives retain metadata (e.g., latitude, longitude, timestamps) and are chunked for efficient loading during training. Each chunk contains 200 timesteps, optimized for sampling speed: about $10~\mathrm{MB}$ per chunk for ERA5 variables and $30~\mathrm{MB}$ for IMERG precipitation. 

Precipitation estimates from model-based products such as ERA5 are generally considered less accurate than those from satellite-based products such as IMERG \cite{seyyedi2015hydrologic, xin2022evaluation}.
To mitigate known artifacts in ERA5 precipitation data---particulary the overestimation of weak precipitation, mis-detection of non-precipitation events \cite{xin2022evaluation}, and unrealistically large values---we clip the precipitation variable using the lowest and highest values from IMERG as thresholds. Very small values are set to zero, while large values are adjusted downward to align with the IMERG data distribution. 

Each variable is then standardized via Z-score normalization using statistics computed across all time steps, latitudes and longitudes, and aggregated across all training regions. The global mean is calculated as the average of region-wise means, and the overall variance is estimated using the pooled variance,
i.e. $$\tilde{\sigma}^2 = \frac{1}{n} \sum_{i=1}^{n} \left( \sigma_i^2 + \mu_i^2 \right) - \left( \frac{1}{n} \sum_{i=1}^{n} \mu_i \right)^2 \text{\,,} $$ where $n$ is the number of training regions, $\mu_{i}$ is the mean, and $\sigma^{2}_{i}$ the variance of region $T_{i}$. 

Following standardization, the two precipitation variables $\text{tp}$ (total precipitation) and $\text{cp}$ (convective precipitation) as well as orography are log-transformed by $\tilde{x} = \log(x \cdot 1000 + 1e5)$. The land-sea mask remains unchanged with values between $0$ and $1$.

For the diffusion model, we additionally perform bilinear interpolation to the ERA5 inputs to match the spatial resolution of IMERG for compatibility with the UNet architecture. Rather than directly predicting the high-resolution target, the UNet is trained to learn the residual between the fine-resolution target and the interpolated coarse-resolution input, following prior work \cite{watt2024generative}. 

\subsubsection*{Temporal splits} We treat every time step as an independent sample. However, as the data is a time series, we split training and evaluation data temporally. The training data in areas $T_1,\ldots,T_{12}$ covers years 2001--2020, the testing data in areas $E_1,\ldots ,E_6 $ covers years 2021--2022.

\subsubsection*{Location splits}\label{loc_splits}
We create RainShift choosing 18 regions worldwide, covering all continents and climate zones, each spanning $20^\circ \times 20^\circ$, as shown in Figure \ref{fig:location_split}. These regions are divided into 12 training regions ($T_1$,\ldots ,$T_{12}$) and six evaluation regions ($E_1$,\ldots ,$E_6$). The six evaluation regions are: Cape Horn, Amazon Basin, West Africa, Horn of Africa, Tibetan Plateau and Melanesia. Training regions are selected from areas with high observational data availability (see Figure \ref{fig:radar} for the example of radar data), while evaluation regions are located in data-scarce areas, primarily in the Global South. 

\subsubsection*{Downscaling task formulation}
The RainShift benchmark focuses on a probabilistic downscaling task, where the goal is to learn the conditional distribution, $p(y|x)$, of high-resolution precipitation, $y$, given a low-resolution forecast and invariant features, $x$.
A generative model, $G$, is trained to approximate the true distribution
$$G(x)\sim p_G(\cdot|x) \text{ such that } p_G(\cdot|x) \approx p(\cdot|x) \text{\,.} $$

Here, the high-resolution 2D target sample $y\in \mathbb{R}^{h_h\times w_h}$ is a single-channel precipitation field from the IMERG satellite data. The input
$x$ consists of both low-resolution and high-resolution components: $x=(x_\ell,x_h)$, with $x_\ell\in\mathbb{R}^{c \times h_\ell\times w_\ell}$ representing a low-resolution $c$-channel forecast (here from ERA5) and $x_h\in \mathbb{R}^{d \times h_h\times w_h}$ denoting $d$-channel high-resolution invariant features (land-sea mask and orography). With an upsampling factor $N\in \mathbb{R}^{+}$, the high-resolution dimensions are given by $h_h=N\cdot h_\ell$ and $w_h=N\cdot w_\ell$. In this benchmark, $N=2.5$ and image patches are square: $ h_\ell=w_\ell=80$ and $h_h=w_h=200$. The channel dimensions are $c=9$ and $d=2$. This downscaling task is illustrated in Figure \ref{fig:setup}.

\subsubsection*{Geographical generalization}
Unlike existing benchmarks that evaluate models within the same geographic region, RainShift is designed to assess generalization across different geographies. We consider $12$ different training regions and $6$ regions for evaluation. Given a training area, $A$, the corresponding local data distribution $p_A$ and samples $(x_A,y_A)\sim p_A$, the goal is to learn the distribution $p_E( \cdot|x_E)$ in a separate evaluation area $E$. Here, training and evaluation areas are disjoint, $A\cap E=\emptyset $. The task is a zero-shot prediction, with no available labels in the evaluation set.
A generative model $G$, is desired to approximate the target distribution
$$G(x)\sim p_G(\cdot|x) \text{ such that } p_G(\cdot|x_E) \approx p_E(\cdot|x_E) \text{\,.} $$

\subsubsection*{Training sub-tasks} \label{sub-tasks}
To simulate different scenarios corresponding to varying levels of data availability, we define four training sub-tasks, as illustrated in Figure \ref{fig:location_split}. Each sub-task is associated with a subset of training areas $A_i\subseteq T=\bigcup_{j=1}^{12} T_j \ \text{for } i=1,\ldots,4$. The subsets are hierarchical---that is, $A_i\subseteq A_{i+1} \ \text{for } i=1,\ldots,3$ reflecting varying levels of observational availability:
\begin{enumerate}
    \item $A_1:=T_1$: a scenario resembling the common task of just training in one geographic area.
    \item $A_2:=T_1\cup T_2 \cup T_5$: inspired by existing works that use both North American and European datasets \cite{cooper2023analysiscgangenerativedeep, prasad2024evaluating}.
    \item $A_3:=A_2 \cup T_{10} \cup T_{11} \cup T_{12}$: an extension of $A_2$ adding three areas with high availability of observational data: two areas in Eastern Asia and one in Eastern Australia.
    \item $A_4:=\bigcup_{i=1}^{12} T_i $: a very optimistic scenario that assumes access to a vast amount of high-resolution observations from a variety of regions and sources, including places with limited data availability.
\end{enumerate}

\subsubsection*{Usage and contribution}
The RainShift benchmark dataset is hosted via Hugging Face at \url{https://huggingface.co/datasets/RainShift/rainshift}. The dataset consists of a zipped Zarr directory per region, resulting in 300 GB of overall data. The download, data-loading, and training instructions can be found in our repository, which will be made available upon acceptance. All baselines are made available via our repository. 

For the purpose of a unified benchmark task, we fix different sets of training and evaluation regions. However, we provide the tools to add new areas of interest to enable optimization or testing in any location worldwide. The instructions to create new regional subsets are provided in our repository, that will be made available upon acceptance.

\subsection*{Baseline models}\label{methods}
We evaluate a diverse set of baseline models to provide a comprehensive view of performance across different model classes. These include a deterministic ResNet model, as well as two successful probabilistic methods: generative adversarial networks (GANs) and a diffusion-based approach. To contextualize these results, we also include bilinear interpolation as a simple baseline to establish a lower performance bound, and in-region training as an upper bound on expected performance.

\subsubsection*{Interpolation baseline}
As precipitation is both an input and output variable, we may construct a simple baseline by bilinearly interpolating ERA5 total precipitation to approximate IMERG data. This helps assess whether poor model performance in specific regions stems from generalization issues or challenges inherent to the input data.

\subsubsection*{ResNet}
Super-resolution CNNs, often ResNet variants, were the first DL tools applied to downscaling \cite{vandal2017}. In deterministic downscaling they still achieve state-of-the-art performance \cite{EnhancingRegionalClimateDownscalingThroughAdvancesinMachineLearning,my_jmlr}. We include a fully-convolutional residual network, following methodology shown to perform highly for precipitation forecast downcaling \cite{Harris2022AGD}. This serves as a deterministic baseline.

\subsubsection*{Generative Adversarial Networks}
GANs are a popular approach for probabilistic downscaling of meteorological data \cite{EnhancingRegionalClimateDownscalingThroughAdvancesinMachineLearning}. In this context, conditional GANs are typically used, where both the generator and the discriminator are conditioned on low-resolution input fields in addition to the generator's noise input. 
We leverage the Wasserstein GAN (WGAN) with gradient penalty \cite{Arjovsky2017WassersteinG}, which is less prone to training instabilities such as mode collapse. The model architecture is based on prior work \cite{Harris2022AGD, leinonen2020stochastic}, and represents a commonly used model in downscaling.

\subsubsection*{Diffusion-based models} 
Diffusion models (DMs), in particular denoising score-matching approaches \cite{pmlr-v37-sohl-dickstein15} are gaining traction in climate- and weather-modeling applications \cite{price_probabilistic_2025, addison2024machinelearningemulationprecipitation}. 
In this work, we follow the diffusion-based downscaling framework introduced in ClimateDiffuse \cite{watt2024generative}, which combines several established components into an effective approach for conditional downscaling of climate fields.
The model is a score-based model trained under the denoising score matching framework, in which a neural network is optimized to learn the score function. This score function is parameterized by a conditional U-Net, conditioned on low-resolution input fields, and the forward and reverse diffusion processes are defined via the stochastic differential equation formulation. Consistent with prior work \cite{watt2024generative}, we also incorporate several key design choices \cite{Karras2022edm}, such as the use of improved preconditioning and a higher-order integration scheme for the differential equation solver.

\subsubsection*{Training details}
The GANs and DMs are trained for 60-168 hours (depending on training area size) with an effective batch size of 128 on 4 NVIDIA A100 GPUs. The ResNets are trained for 45-280 hours on 2 NVIDIA RTX8000. The years 2019 and 2020 are used as validation data for hyperparameter tuning and choosing the best checkpoint. 

\subsection*{Evaluation}
To evaluate spatial generalization ability of the models, we compare their performance across the different training scenarios using the continuous ranked probability score as a metric. Model performance is reported as absolute scores (see Table \ref{tab:crps_results}), relative improvements over interpolation baseline (see Figure \ref{fig:rel_bilinear}) and relative performance compared to training directly on the target area (see Figure \ref{fig:rel_target}). 

\subsubsection*{Quantitative evaluation} \label{evaluation}
The overall most successful model for each subtask (defined by $A_i\ \text{for } i=1,\ldots ,4$) is determined by the point-wise continuous ranked probability score (CRPS) \cite{Broecker2007}, calculated using $8$ samples at each of the six target locations. The CRPS is a metric used to evaluate the accuracy of probabilistic forecasts. For a given forecast probability distribution $F$ and the observed outcome ${y}$, the CRPS metric is defined as follows:
    \begin{equation}\label{eq:crps_definition}
 \text{CRPS}(F, {y}) = \int_{-\infty}^{\infty} [F(z) - \mathbf{1}(z \geq {y})]^2 \text{dz} \text{\,.}
\end{equation}
Here, $F(z)$ is the cumulative distribution function of the forecast distribution at point $z$ and
$\bf{1}(\cdot)$ the indicator function. For a deterministic forecast, the CRPS reduces to the mean absolute error. 

\subsubsection*{In-area training}
The standard evaluation in deep learning-based downscaling consists of training and evaluating in the same area. To show that the geographical generalization is indeed a challenge we also report scores that use the target area labels. For this, we train the above mentioned models, ResNets, GANs and DMs on the respective target areas $E_1,\ldots,E_6$ directly, while keeping the temporal train-test split: training on years 2001-2020 and testing on years 2021 and 2022. 

\begin{table*}[htb] 
\vskip 0.15in
\begin{center}
\begin{small}
\begin{tabular}{lccccccc}
\toprule
&  &\multicolumn{6}{c}{Target Areas}\\
 &   & E1 & E2 & E3 & E4 & E5 & E6 \\
  Model  & Training  & Cape  & Amazon & West  & Horn & Tibetan  & Melanesia\\
  & Areas  &  Horn & Basin & Africa & of Africa &  Plateau & \\
  \toprule
ERA5 Interp. & - & 0.120& 0.258 & 0.113 & 0.084 & 0.050 & 0.434 \\ 
 \toprule
GAN & $A_1$ &  \textbf{0.075} & 0.179 &0.093 &0.055 & \textbf{0.027} & 0.326 \\
DM & $A_1$ &   0.091 & \textbf{0.173} & \textbf{0.084} & \textbf{0.054} & 0.028 & \textbf{0.295}\\
ResNet & $A_1$ & 0.083& 0.187 & 0.308 & 0.057 & 18113.558 & 0.389\\
\midrule
GAN & $A_2$ & \textbf{0.076} & 0.178 & 0.088 & 0.057 &0.032 &0.324 \\
DM & $A_2$ &  0.078 & \textbf{0.177} & \textbf{0.080} & 0.056 & \textbf{0.027} & \textbf{0.288}\\
ResNet & $A_2$ & 0.083 & 0.181 & \textbf{0.080} & \textbf{0.055} & 0.034 & 0.315\\
\midrule
GAN & $A_3$ & \textbf{0.072}& \textbf{0.169}& 0.080& \textbf{0.053}& \textbf{0.024}& \textbf{0.288} \\
DM & $A_3$ &  0.074 & 0.171 & 0.079 & 0.054 & 0.025 & 0.290 \\
ResNet & $A_3$ &0.081 &0.181 & 0.080 & 0.055 &0.025&0.306 \\
\midrule
GAN & $A_4$ & \textbf{0.070} &\textbf{0.167}& \textbf{0.077}& \textbf{0.054}& 0.025& \textbf{0.272}\\
DM & $A_4$ &  0.072 & 0.170 & \textbf{0.077} & 0.056 & 0.027 & 0.284\\
ResNet & $A_4$ &  0.083& 0.179& 0.078 &0.055 & \textbf{0.024} & 0.298 \\
\toprule
GAN & $E_i$ & 0.077 & 0.180 & \textcolor{blue}{0.072} 
& \textcolor{blue}{0.053} & 0.025& \textcolor{blue}{0.272}\\
DM & $ E_i$ & \textcolor{blue}{0.070}& \textcolor{blue}{0.171} & \textcolor{blue}{0.075} & \textcolor{blue}{0.054} & \textcolor{blue}{0.023} & \textcolor{blue}{0.275} \\
ResNet & $ E_i$ &  \textcolor{blue}{0.078} & 0.181 & \textcolor{blue}{0.078} & 0.057 & 0.025 & 0.311\\
\bottomrule
\end{tabular}
\end{small}
\end{center}
\caption{\textbf{Accuracy of models on spatial generalization tasks.}
Test CRPS (lower better) for precipitation in mm/h on the designated evaluation area and averaged over test years $2021 - 2022$. Shown is the mean, pixel-wise CRPS of 8 ensemble members. For deterministic models (ResNet and bilinear interpolation of ERA5 precipitation data), the MAE is shown. We report the mean precipitation amount in input and target data in mm/h. Best scores per subtask are in bold. The last three rows with scores in blue are not contestants in the benchmark but show what is possible when training directly on the target.}
\vskip -0.1in
\label{tab:crps_results}
\end{table*}

\subsubsection*{Qualitative evaluation} 
\begin{figure}[htb]
\begin{center}
\vskip 0.1in
\centerline{\includegraphics[width=10cm]{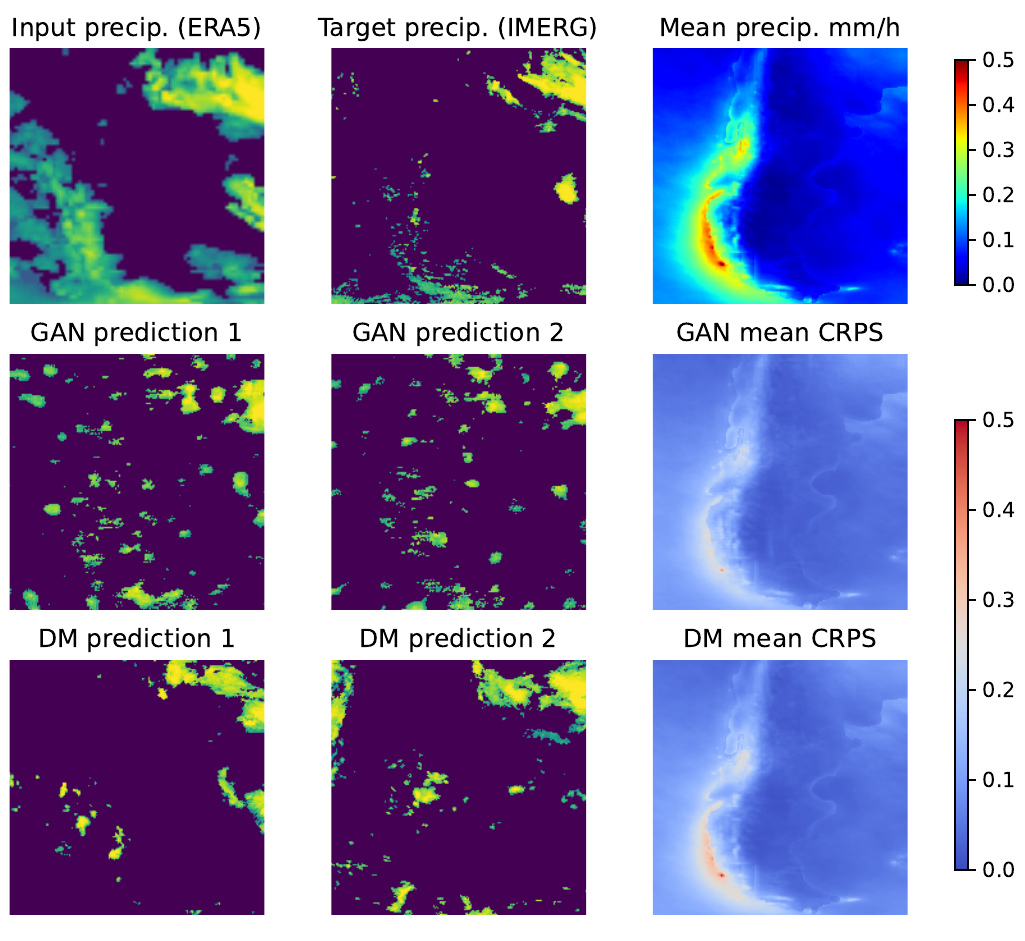}}
\caption{\textbf{Qualitative comparison of downscaled precipitation fields.} This plot shows a sample, one time step from the evaluation set in the Cape Horn area in the first two columns and aggregated features in the last column. The random sample includes the matching input (ERA5) and target (IMERG) precipitation values in logarithmic scale and two samples each from the two generative approaches, GANs and DMs. The right column shows on top, the mean precipitation over the whole testing period (2021-2022) as well as the spatial distribution of CRPS scores for GAN and DM.}
\label{samples}
\end{center}
\vskip -0.2in
\end{figure}

In addition to CRPS scores, we provide a qualitative comparison of downscaled precipitation fields for one sample (see Figure \ref{samples}) as well as temporally averaged precipitation and CRPS values.

\subsection*{Background}
Model generalization to new geographic regions is an active research area in applied machine learning, particularly in remote sensing and biodiversity modeling. In agricultural classification and segmentation, approaches such as task-informed meta-learning \cite{tseng2022timltaskinformedmetalearningagriculture}, versions of model-agnostic meta-learning \cite{rußwurm2020metalearning}, and multi-source unsupervised domain adaptation \cite{Wang31122022} have shown promise in adapting to new regions with minimal data. In biodiversity monitoring, previous work integrates remote sensing and citizen science data to improve generalization in data-sparse regions like Kenya \cite{NEURIPS2023_ef7653bb}, while spatial implicit neural representations have been leveraged for scalable global species range estimation using noisy, sparse data \cite{SINR_icml23}.

In climate science, spatial generalization in deep learning-based downscaling remains underexplored. A few recent studies have tested model transferability between subregions \cite{10440025}. For instance, some works examine generalization between different areas on the US West Coast \cite{Sha2020DeepLearningBasedGD, DeepLearningBasedGriddedDownscalingofSurfaceMeteorologicalVariablesinComplexTerrainPartIIDailyPrecipitation}. Others analyze performance across regions in the UK and the United States \cite{cooper2023analysiscgangenerativedeep} or evaluate generalization from the DACH region (Germany, Austria, and Switzerland) to North America \cite{prasad2024evaluatingtransferabilitypotentialdeep}. While these efforts provide valuable insights, their geographic scope remains limited. 

Another important development has been the creation of benchmark datasets aimed at standardizing and accelerating machine learning methods for Earth System Modeling.
Notable examples include WeatherBench \cite{rasp2023weatherbench}, ClimateBench \cite{WatsonParris2022ClimateBenchVA}, ClimateLearn \cite{Nguyen2023ClimateLearnBM} and ClimateSet \cite{Kaltenborn2023ClimateSetAL} which have provided standardized datasets, tasks and evaluation frameworks. 
Several benchmarks specifically target precipitation forecasting and downscaling. RainBench introduces a global benchmark for precipitation forecasting based on IMERG data \cite{Martini_Kalaitzis_Chantry_Watson-Parris_Bilinski_2021}. RainNet focuses on precipitation super-resolution, targeting a region on the US East Coast with single-variable input data in a deterministic setup \cite{Chen2020RainNetAL}. The ClimateLearn benchmark supports evaluation of downscaling techniques that map low-resolution CMIP6 model outputs to high-reslution ERA5 data \cite{Nguyen2023ClimateLearnBM}. These efforts have made climate modeling more accessible to the broader machine learning community. Despite this, no such existing benchmark has yet specifically addressed the challenge of generalizing across geographies. In this paper, we present RainShift, seeking to fill this gap by introducing a large-scale global benchmark dataset specifically designed to evaluate and improve the geographical generalization of deep learning-based downscaling.

\subsection*{Quantile mapping for geographical generalization}
Geographic generalization is a central challenge in statistical downscaling, resulting from differences in climatic conditions and their underlying processes across geographical areas. Such differences can lead to substantial performance drops when models trained in one region are applied to new regions with distinct climatic characteristics. In our experiments, we observe large performance drops in target regions whose precipitation distributions differ markedly from those of the training regions. Therefore, a promising strategy to address these distributional shifts is to align the input data distributions of the training and target regions. To this end, we use a strategy based on quantile mapping.

\begin{figure}[htb]
\begin{center}
\vskip 0.1in
\centerline{\includegraphics[width=\columnwidth]{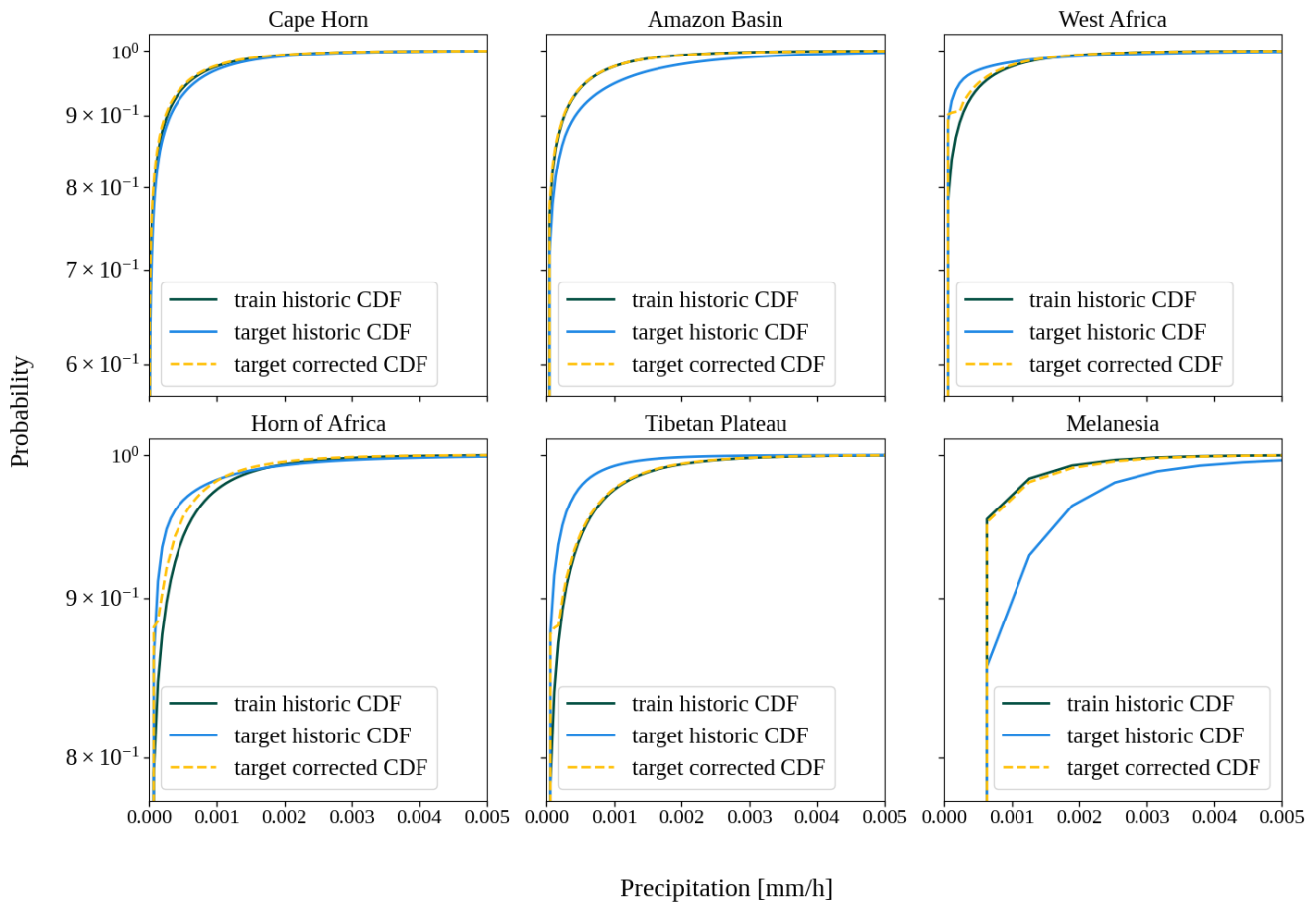}}
\caption{\textbf{Cumulative distribution functions (CDFs) of precipitation inputs for training and target regions.} For each target region, a mapping is constructed between the historic CDF of the training region and the historic precipitation inputs of the target region. This mapping is subsequently used to correct the future input data from the target region, aligning them more closely with the training data distribution. Since precipitation is highly right-skewed with many zero and near-zero values, which causes most values to fall within the first bin of the CDF, we show the logarithm of the CDF to better visualize the discrepancies in distributions.}
\label{cdf}
\end{center}
\vskip -0.2in
\end{figure}

While the quantile mapping (QM) technique is traditionally used to correct systematic distributional biases in climate model simulations relative to observations (leveraging historical relationships between simulations and observations to adjust future simulations) \cite{cannon2015bias}, we apply it to address mismatches between the input distributions of the training and target regions. By aligning the input distribution of the target region more closely with that of the training data, the model may be better able to generalize and generate more reliable predictions in unseen regions. Specifically, we learn a mapping between the cumulative distribution functions (CDFs) of the precipitation input data of the training regions, $F_{\text{train}, h}$, and that of the target regions, $F_{\text{target}, h}$, over a historical period $h$. This results in the following transfer function:
$$
\hat{x}_{\text{target}, f}(t) = F^{-1}_{\text{train}, h} \left( F_{\text{target}, h} \left[ x_{\text{target}, f}(t) \right] \right) \text{\,,}
$$
which adapts the precipitation value $x_{\text{target}, f}(t)$ from a target region within some future period $f$. For a non-negative variable such as precipitation we use the multiplicative variant of quantile mapping \cite{cannon2015bias}, where the values are lower bounded by zero. 
We perform quantile mapping to the low-resolution precipitation inputs using $1000$ quantiles prior to data normalization during inference. Figure \ref{cdf} illustrates the discrepancy between the CDFs of the training and target regions, which is pronounced for all target regions except Cape Horn. After correction, the target region CDFs align more closely with the training CDF. Consistent with this reduction in distributional mismatch, we observe in Table \ref{tab:qm_results} that the ResNet model evaluated with quantile-based aligned inputs achieves improved performance in all target regions except Cape Horn, compared to using unaligned inputs.

\section*{Data availability}
The RainShift benchmark dataset is hosted via Hugging Face at \url{https://huggingface.co/datasets/RainShift/rainshift}. The dataset consist of a zipped Zarr directory per region, resulting in overall 300GB of data. The download, data-loading, and training instructions can be found in our repository, that will be made available upon acceptance.




\bibliography{sample}

\begin{thebibliography}{10}
\urlstyle{rm}
\expandafter\ifx\csname url\endcsname\relax
  \def\url#1{\texttt{#1}}\fi
\expandafter\ifx\csname urlprefix\endcsname\relax\def\urlprefix{URL }\fi
\expandafter\ifx\csname doiprefix\endcsname\relax\def\doiprefix{DOI: }\fi
\providecommand{\bibinfo}[2]{#2}
\providecommand{\eprint}[2][]{\url{#2}}

\bibitem{seneviratne2021weather}
\bibinfo{author}{Seneviratne, S.~I.} \emph{et~al.}
\newblock \bibinfo{title}{Weather and climate extreme events in a changing climate}.
\newblock In \emph{\bibinfo{booktitle}{Climate Change 2021: The Physical Science Basis. Contribution of Working Group I to the Sixth Assessment Report of the Intergovernmental Panel on Climate Change}} (\bibinfo{publisher}{Cambridge University Press}, \bibinfo{year}{2021}).

\bibitem{ombadi2023warming}
\bibinfo{author}{Ombadi, M.}, \bibinfo{author}{Risser, M.~D.}, \bibinfo{author}{Rhoades, A.~M.} \& \bibinfo{author}{Varadharajan, C.}
\newblock \bibinfo{journal}{\bibinfo{title}{A warming-induced reduction in snow fraction amplifies rainfall extremes}}.
\newblock {\emph{\JournalTitle{Nature}}} \textbf{\bibinfo{volume}{619}}, \bibinfo{pages}{305--310} (\bibinfo{year}{2023}).

\bibitem{xiong2024climate}
\bibinfo{author}{Xiong, J.} \& \bibinfo{author}{Yang, Y.}
\newblock \bibinfo{journal}{\bibinfo{title}{Climate change and hydrological extremes}}.
\newblock {\emph{\JournalTitle{Current Climate Change Reports}}} \textbf{\bibinfo{volume}{11}}, \bibinfo{pages}{1} (\bibinfo{year}{2024}).

\bibitem{maraun2010precipitation}
\bibinfo{author}{Maraun, D.} \emph{et~al.}
\newblock \bibinfo{journal}{\bibinfo{title}{Precipitation downscaling under climate change: Recent developments to bridge the gap between dynamical models and the end user}}.
\newblock {\emph{\JournalTitle{Reviews of geophysics}}} \textbf{\bibinfo{volume}{48}} (\bibinfo{year}{2010}).

\bibitem{pendergrass2017precipitation}
\bibinfo{author}{Pendergrass, A.~G.}, \bibinfo{author}{Knutti, R.}, \bibinfo{author}{Lehner, F.}, \bibinfo{author}{Deser, C.} \& \bibinfo{author}{Sanderson, B.~M.}
\newblock \bibinfo{journal}{\bibinfo{title}{Precipitation variability increases in a warmer climate}}.
\newblock {\emph{\JournalTitle{Scientific reports}}} \textbf{\bibinfo{volume}{7}}, \bibinfo{pages}{17966} (\bibinfo{year}{2017}).

\bibitem{fosser2024convection}
\bibinfo{author}{Fosser, G.} \emph{et~al.}
\newblock \bibinfo{journal}{\bibinfo{title}{Convection-permitting climate models offer more certain extreme rainfall projections}}.
\newblock {\emph{\JournalTitle{NPJ Climate and atmospheric science}}} \textbf{\bibinfo{volume}{7}}, \bibinfo{pages}{51} (\bibinfo{year}{2024}).

\bibitem{vandal2017}
\bibinfo{author}{Vandal, T.} \emph{et~al.}
\newblock \bibinfo{journal}{\bibinfo{title}{Deepsd: Generating high resolution climate change projections through single image super-resolution}}.
\newblock {\emph{\JournalTitle{Association for Computing Machinery}}} \bibinfo{pages}{1663–1672}, \doiprefix\url{10.1145/3097983.3098004} (\bibinfo{year}{2017}).

\bibitem{Sha2020DeepLearningBasedGD}
\bibinfo{author}{Sha, Y.}, \bibinfo{author}{Gagne, D.~J.}, \bibinfo{author}{West, G.} \& \bibinfo{author}{Stull, R.}
\newblock \bibinfo{journal}{\bibinfo{title}{Deep-learning-based gridded downscaling of surface meteorological variables in complex terrain. part i: Daily maximum and minimum 2-m temperature}}.
\newblock {\emph{\JournalTitle{Journal of Applied Meteorology and Climatology}}}  (\bibinfo{year}{2020}).

\bibitem{https://doi.org/10.1002/met.1961}
\bibinfo{author}{Höhlein, K.}, \bibinfo{author}{Kern, M.}, \bibinfo{author}{Hewson, T.} \& \bibinfo{author}{Westermann, R.}
\newblock \bibinfo{journal}{\bibinfo{title}{A comparative study of convolutional neural network models for wind field downscaling}}.
\newblock {\emph{\JournalTitle{Meteorological Applications}}} \textbf{\bibinfo{volume}{27}}, \bibinfo{pages}{e1961}, \doiprefix\url{https://doi.org/10.1002/met.1961} (\bibinfo{year}{2020}).
\newblock \eprint{https://rmets.onlinelibrary.wiley.com/doi/pdf/10.1002/met.1961}.

\bibitem{10.1145/3394486.3403366}
\bibinfo{author}{Liu, Y.}, \bibinfo{author}{Ganguly, A.~R.} \& \bibinfo{author}{Dy, J.}
\newblock \bibinfo{title}{Climate downscaling using ynet: A deep convolutional network with skip connections and fusion}.
\newblock In \emph{\bibinfo{booktitle}{Proceedings of the 26th ACM SIGKDD International Conference on Knowledge Discovery \& Data Mining}}, KDD '20, \bibinfo{pages}{3145–3153}, \doiprefix\url{10.1145/3394486.3403366} (\bibinfo{publisher}{Association for Computing Machinery}, \bibinfo{address}{New York, NY, USA}, \bibinfo{year}{2020}).

\bibitem{8588749}
\bibinfo{author}{Rocha~Rodrigues, E.}, \bibinfo{author}{Oliveira, I.}, \bibinfo{author}{Cunha, R.} \& \bibinfo{author}{Netto, M.}
\newblock \bibinfo{title}{Deepdownscale: A deep learning strategy for high-resolution weather forecast}.
\newblock In \emph{\bibinfo{booktitle}{2018 IEEE 14th International Conference on e-Science (e-Science)}}, \bibinfo{pages}{415--422}, \doiprefix\url{10.1109/eScience.2018.00130} (\bibinfo{year}{2018}).

\bibitem{Goodfellow2014GenerativeAN}
\bibinfo{author}{Goodfellow, I.~J.} \emph{et~al.}
\newblock \bibinfo{journal}{\bibinfo{title}{Generative adversarial networks}}.
\newblock {\emph{\JournalTitle{Communications of the ACM}}} \textbf{\bibinfo{volume}{63}}, \bibinfo{pages}{139 -- 144} (\bibinfo{year}{2014}).

\bibitem{Harris2022AGD}
\bibinfo{author}{Harris, L.}, \bibinfo{author}{McRae, A. T.~T.}, \bibinfo{author}{Chantry, M.}, \bibinfo{author}{Dueben, P.~D.} \& \bibinfo{author}{Palmer, T.~N.}
\newblock \bibinfo{journal}{\bibinfo{title}{A generative deep learning approach to stochastic downscaling of precipitation forecasts}}.
\newblock {\emph{\JournalTitle{Journal of Advances in Modeling Earth Systems}}} \textbf{\bibinfo{volume}{14}} (\bibinfo{year}{2022}).

\bibitem{cooper2023analysiscgangenerativedeep}
\bibinfo{author}{Cooper, F.~C.}, \bibinfo{author}{McRae, A. T.~T.}, \bibinfo{author}{Chantry, M.}, \bibinfo{author}{Antonio, B.} \& \bibinfo{author}{Palmer, T.~N.}
\newblock \bibinfo{title}{Further analysis of cgan: A system for generative deep learning post-processing of precipitation} (\bibinfo{year}{2023}).
\newblock \eprint{2309.15689}.

\bibitem{SohlDickstein2015DeepUL}
\bibinfo{author}{Sohl-Dickstein, J.~N.}, \bibinfo{author}{Weiss, E.~A.}, \bibinfo{author}{Maheswaranathan, N.} \& \bibinfo{author}{Ganguli, S.}
\newblock \bibinfo{journal}{\bibinfo{title}{Deep unsupervised learning using nonequilibrium thermodynamics}}.
\newblock {\emph{\JournalTitle{ArXiv}}} \textbf{\bibinfo{volume}{abs/1503.03585}} (\bibinfo{year}{2015}).

\bibitem{Mardani2023GenerativeRD}
\bibinfo{author}{Mardani, M.} \emph{et~al.}
\newblock \bibinfo{journal}{\bibinfo{title}{Generative residual diffusion modeling for km-scale atmospheric downscaling}}.
\newblock {\emph{\JournalTitle{ArXiv}}} \textbf{\bibinfo{volume}{abs/2309.15214}} (\bibinfo{year}{2023}).

\bibitem{Wan2023DebiasCS}
\bibinfo{author}{Wan, Z.~Y.} \emph{et~al.}
\newblock \bibinfo{journal}{\bibinfo{title}{Debias coarsely, sample conditionally: Statistical downscaling through optimal transport and probabilistic diffusion models}}.
\newblock {\emph{\JournalTitle{ArXiv}}} \textbf{\bibinfo{volume}{abs/2305.15618}} (\bibinfo{year}{2023}).

\bibitem{addison2024machinelearningemulationprecipitation}
\bibinfo{author}{Addison, H.}, \bibinfo{author}{Kendon, E.}, \bibinfo{author}{Ravuri, S.}, \bibinfo{author}{Aitchison, L.} \& \bibinfo{author}{Watson, P.~A.}
\newblock \bibinfo{title}{Machine learning emulation of precipitation from km-scale regional climate simulations using a diffusion model} (\bibinfo{year}{2024}).
\newblock \eprint{2407.14158}.

\bibitem{ling_diffusion_2024}
\bibinfo{author}{Ling, F.}, \bibinfo{author}{Lu, Z.}, \bibinfo{author}{Luo, J.~J.} \emph{et~al.}
\newblock \bibinfo{journal}{\bibinfo{title}{{Diffusion model-based probabilistic downscaling for 180-year East Asian climate reconstruction}}}.
\newblock {\emph{\JournalTitle{npj Climate and Atmospheric Science}}} \textbf{\bibinfo{volume}{7}}, \bibinfo{pages}{131}, \doiprefix\url{10.1038/s41612-024-00679-1} (\bibinfo{year}{2024}).

\bibitem{tropical_globe_radar_database}
\bibinfo{author}{{Tropical Globe}}.
\newblock \bibinfo{title}{Tropical globe radar database}.
\newblock \bibinfo{howpublished}{\url{https://tropicalglobe.com/radar_database/}} (\bibinfo{year}{2025}).
\newblock \bibinfo{note}{Accessed: 2025-01-26}.

\bibitem{Arjovsky2017WassersteinG}
\bibinfo{author}{Arjovsky, M.}, \bibinfo{author}{Chintala, S.} \& \bibinfo{author}{Bottou, L.}
\newblock \bibinfo{journal}{\bibinfo{title}{Wasserstein {GAN}}}.
\newblock {\emph{\JournalTitle{ArXiv}}} \textbf{\bibinfo{volume}{abs/1701.07875}} (\bibinfo{year}{2017}).

\bibitem{watt2024generative}
\bibinfo{author}{Watt, R.~A.} \& \bibinfo{author}{Mansfield, L.~A.}
\newblock \bibinfo{title}{Generative diffusion-based downscaling for climate} (\bibinfo{year}{2024}).
\newblock \eprint{2404.17752}.

\bibitem{prasad2024evaluating}
\bibinfo{author}{Prasad, A.} \emph{et~al.}
\newblock \bibinfo{journal}{\bibinfo{title}{Evaluating the transferability potential of deep learning models for climate downscaling}}.
\newblock {\emph{\JournalTitle{ICML Workshop Machine Learning for Earth System Modeling}}}  (\bibinfo{year}{2024}).

\bibitem{Klemmer2023SatCLIPGG}
\bibinfo{author}{Klemmer, K.}, \bibinfo{author}{Rolf, E.}, \bibinfo{author}{Robinson, C.}, \bibinfo{author}{Mackey, L.} \& \bibinfo{author}{Ru{\ss}wurm, M.}
\newblock \bibinfo{journal}{\bibinfo{title}{Satclip: Global, general-purpose location embeddings with satellite imagery}}.
\newblock {\emph{\JournalTitle{ArXiv}}} \textbf{\bibinfo{volume}{abs/2311.17179}} (\bibinfo{year}{2023}).

\bibitem{Wang31122022}
\bibinfo{author}{Wang, Y.} \emph{et~al.}
\newblock \bibinfo{journal}{\bibinfo{title}{Exploring the potential of multi-source unsupervised domain adaptation in crop mapping using sentinel-2 images}}.
\newblock {\emph{\JournalTitle{GIScience \& Remote Sensing}}} \textbf{\bibinfo{volume}{59}}, \bibinfo{pages}{2247--2265}, \doiprefix\url{10.1080/15481603.2022.2156123} (\bibinfo{year}{2022}).
\newblock \eprint{https://doi.org/10.1080/15481603.2022.2156123}.

\bibitem{rußwurm2020metalearning}
\bibinfo{author}{Rußwurm, M.}, \bibinfo{author}{Wang, S.}, \bibinfo{author}{Körner, M.} \& \bibinfo{author}{Lobell, D.}
\newblock \bibinfo{journal}{\bibinfo{title}{Meta-learning for few-shot land cover classification}}.
\newblock {\emph{\JournalTitle{Preprint arXiv 2004.13390}}}  (\bibinfo{year}{2020}).

\bibitem{harder2023hard}
\bibinfo{author}{Harder, P.} \emph{et~al.}
\newblock \bibinfo{journal}{\bibinfo{title}{Hard-constrained deep learning for climate downscaling}}.
\newblock {\emph{\JournalTitle{Journal of Machine Learning Research}}} \textbf{\bibinfo{volume}{24}}, \bibinfo{pages}{1--40} (\bibinfo{year}{2023}).

\bibitem{era5}
\bibinfo{author}{Hersbach, H.} \emph{et~al.}
\newblock \bibinfo{journal}{\bibinfo{title}{The era5 global reanalysis}}.
\newblock {\emph{\JournalTitle{Quarterly Journal of the Royal Meteorological Society}}} \textbf{\bibinfo{volume}{146}}, \bibinfo{pages}{1999--2049}, \doiprefix\url{https://doi.org/10.1002/qj.3803} (\bibinfo{year}{2020}).

\bibitem{Hewson2020ALP}
\bibinfo{author}{Hewson, T.} \& \bibinfo{author}{Pillosu, F.~M.}
\newblock \bibinfo{journal}{\bibinfo{title}{A low-cost post-processing technique improves weather forecasts around the world}}.
\newblock {\emph{\JournalTitle{Communications Earth \& Environment}}} \textbf{\bibinfo{volume}{2}} (\bibinfo{year}{2020}).

\bibitem{huffman_imerg_2014}
\bibinfo{author}{Huffman, G.} \emph{et~al.}
\newblock \bibinfo{title}{{Integrated Multi-satellitE Retrievals for GPM (IMERG), version 4.4}}.
\newblock \bibinfo{howpublished}{{NASA's Precipitation Processing Center}} (\bibinfo{year}{2014}).
\newblock \bibinfo{note}{Accessed: 31 March, 2015}.

\bibitem{huffman_imerg_v07}
\bibinfo{author}{Huffman, G.~J.} \emph{et~al.}
\newblock \bibinfo{title}{{IMERG V07 Release Notes}}.
\newblock \bibinfo{howpublished}{\url{https://gpm.nasa.gov/resources/documents/imerg-v07-release-notes}} (\bibinfo{year}{2024}).
\newblock \bibinfo{note}{Accessed: 2025-01-26}.

\bibitem{alistair_miles_2020_3773450}
\bibinfo{author}{Miles, A.} \emph{et~al.}
\newblock \bibinfo{title}{zarr-developers/zarr-python: v2.4.0}, \doiprefix\url{10.5281/zenodo.3773450} (\bibinfo{year}{2020}).

\bibitem{seyyedi2015hydrologic}
\bibinfo{author}{Seyyedi, H.}, \bibinfo{author}{Anagnostou, E.~N.}, \bibinfo{author}{Beighley, E.} \& \bibinfo{author}{McCollum, J.}
\newblock \bibinfo{journal}{\bibinfo{title}{Hydrologic evaluation of satellite and reanalysis precipitation datasets over a mid-latitude basin}}.
\newblock {\emph{\JournalTitle{Atmospheric Research}}} \textbf{\bibinfo{volume}{164}}, \bibinfo{pages}{37--48} (\bibinfo{year}{2015}).

\bibitem{xin2022evaluation}
\bibinfo{author}{Xin, Y.} \emph{et~al.}
\newblock \bibinfo{journal}{\bibinfo{title}{Evaluation of imerg and era5 precipitation products over the mongolian plateau}}.
\newblock {\emph{\JournalTitle{Scientific reports}}} \textbf{\bibinfo{volume}{12}}, \bibinfo{pages}{21776} (\bibinfo{year}{2022}).

\bibitem{EnhancingRegionalClimateDownscalingThroughAdvancesinMachineLearning}
\bibinfo{author}{Rampal, N.} \emph{et~al.}
\newblock \bibinfo{journal}{\bibinfo{title}{Enhancing regional climate downscaling through advances in machine learning}}.
\newblock {\emph{\JournalTitle{Artificial Intelligence for the Earth Systems}}} \doiprefix\url{10.1175/AIES-D-23-0066.1} (\bibinfo{year}{2024}).

\bibitem{my_jmlr}
\bibinfo{author}{Harder, P.} \emph{et~al.}
\newblock \bibinfo{journal}{\bibinfo{title}{Hard-constrained deep learning for climate downscaling}}.
\newblock {\emph{\JournalTitle{Journal of Machine Learning Research}}} \textbf{\bibinfo{volume}{24}}, \bibinfo{pages}{1--40} (\bibinfo{year}{2023}).

\bibitem{leinonen2020stochastic}
\bibinfo{author}{Leinonen, J.}, \bibinfo{author}{Nerini, D.} \& \bibinfo{author}{Berne, A.}
\newblock \bibinfo{journal}{\bibinfo{title}{Stochastic super-resolution for downscaling time-evolving atmospheric fields with a generative adversarial network}}.
\newblock {\emph{\JournalTitle{IEEE Transactions on Geoscience and Remote Sensing}}} \textbf{\bibinfo{volume}{59}}, \bibinfo{pages}{7211--7223} (\bibinfo{year}{2020}).

\bibitem{pmlr-v37-sohl-dickstein15}
\bibinfo{author}{Sohl-Dickstein, J.}, \bibinfo{author}{Weiss, E.}, \bibinfo{author}{Maheswaranathan, N.} \& \bibinfo{author}{Ganguli, S.}
\newblock \bibinfo{title}{Deep unsupervised learning using nonequilibrium thermodynamics}.
\newblock In \bibinfo{editor}{Bach, F.} \& \bibinfo{editor}{Blei, D.} (eds.) \emph{\bibinfo{booktitle}{Proceedings of the 32nd International Conference on Machine Learning}}, vol.~\bibinfo{volume}{37} of \emph{\bibinfo{series}{Proceedings of Machine Learning Research}}, \bibinfo{pages}{2256--2265} (\bibinfo{publisher}{PMLR}, \bibinfo{address}{Lille, France}, \bibinfo{year}{2015}).

\bibitem{price_probabilistic_2025}
\bibinfo{author}{Price, I.}, \bibinfo{author}{Sanchez-Gonzalez, A.}, \bibinfo{author}{Alet, F.} \emph{et~al.}
\newblock \bibinfo{journal}{\bibinfo{title}{Probabilistic weather forecasting with machine learning}}.
\newblock {\emph{\JournalTitle{Nature}}} \textbf{\bibinfo{volume}{637}}, \bibinfo{pages}{84--90}, \doiprefix\url{10.1038/s41586-024-08252-9} (\bibinfo{year}{2025}).

\bibitem{Karras2022edm}
\bibinfo{author}{Karras, T.}, \bibinfo{author}{Aittala, M.}, \bibinfo{author}{Aila, T.} \& \bibinfo{author}{Laine, S.}
\newblock \bibinfo{title}{Elucidating the design space of diffusion-based generative models}.
\newblock In \emph{\bibinfo{booktitle}{Proc. NeurIPS}} (\bibinfo{year}{2022}).

\bibitem{Broecker2007}
\bibinfo{author}{Broecker, J.} \& \bibinfo{author}{Smith, L.~A.}
\newblock \emph{\bibinfo{title}{Increasing the Reliability of Reliability Diagrams}}, vol.~\bibinfo{volume}{22} (\bibinfo{publisher}{Weather and Forecasting}, \bibinfo{year}{2007}).

\bibitem{tseng2022timltaskinformedmetalearningagriculture}
\bibinfo{author}{Tseng, G.}, \bibinfo{author}{Kerner, H.} \& \bibinfo{author}{Rolnick, D.}
\newblock \bibinfo{title}{Timl: Task-informed meta-learning for agriculture} (\bibinfo{year}{2022}).
\newblock \eprint{2202.02124}.

\bibitem{NEURIPS2023_ef7653bb}
\bibinfo{author}{Teng, M.} \emph{et~al.}
\newblock \bibinfo{title}{Satbird: a dataset for bird species distribution modeling using remote sensing and citizen science data}.
\newblock In \bibinfo{editor}{Oh, A.} \emph{et~al.} (eds.) \emph{\bibinfo{booktitle}{Advances in Neural Information Processing Systems}}, vol.~\bibinfo{volume}{36}, \bibinfo{pages}{75925--75950} (\bibinfo{publisher}{Curran Associates, Inc.}, \bibinfo{year}{2023}).

\bibitem{SINR_icml23}
\bibinfo{author}{Cole, E.} \emph{et~al.}
\newblock \bibinfo{title}{{Spatial Implicit Neural Representations for Global-Scale Species Mapping}}.
\newblock In \emph{\bibinfo{booktitle}{ICML}} (\bibinfo{year}{2023}).

\bibitem{10440025}
\bibinfo{author}{Zhu, H.} \& \bibinfo{author}{Zhou, Q.}
\newblock \bibinfo{journal}{\bibinfo{title}{Advancing satellite-derived precipitation downscaling in data-sparse area through deep transfer learning}}.
\newblock {\emph{\JournalTitle{IEEE Transactions on Geoscience and Remote Sensing}}} \textbf{\bibinfo{volume}{62}}, \bibinfo{pages}{1--13}, \doiprefix\url{10.1109/TGRS.2024.3367332} (\bibinfo{year}{2024}).

\bibitem{DeepLearningBasedGriddedDownscalingofSurfaceMeteorologicalVariablesinComplexTerrainPartIIDailyPrecipitation}
\bibinfo{author}{Sha, Y.}, \bibinfo{author}{II, D. J.~G.}, \bibinfo{author}{West, G.} \& \bibinfo{author}{Stull, R.}
\newblock \bibinfo{journal}{\bibinfo{title}{Deep-learning-based gridded downscaling of surface meteorological variables in complex terrain. part ii: Daily precipitation}}.
\newblock {\emph{\JournalTitle{Journal of Applied Meteorology and Climatology}}} \textbf{\bibinfo{volume}{59}}, \bibinfo{pages}{2075 -- 2092}, \doiprefix\url{10.1175/JAMC-D-20-0058.1} (\bibinfo{year}{2020}).

\bibitem{prasad2024evaluatingtransferabilitypotentialdeep}
\bibinfo{author}{Prasad, A.} \emph{et~al.}
\newblock \bibinfo{title}{Evaluating the transferability potential of deep learning models for climate downscaling} (\bibinfo{year}{2024}).
\newblock \eprint{2407.12517}.

\bibitem{rasp2023weatherbench}
\bibinfo{author}{Rasp, S.} \emph{et~al.}
\newblock \bibinfo{title}{Weatherbench 2: A benchmark for the next generation of data-driven global weather models} (\bibinfo{year}{2023}).
\newblock \eprint{2308.15560}.

\bibitem{WatsonParris2022ClimateBenchVA}
\bibinfo{author}{Watson‐Parris, D.} \emph{et~al.}
\newblock \bibinfo{journal}{\bibinfo{title}{Climatebench v1.0: A benchmark for data‐driven climate projections}}.
\newblock {\emph{\JournalTitle{Journal of Advances in Modeling Earth Systems}}} \textbf{\bibinfo{volume}{14}} (\bibinfo{year}{2022}).

\bibitem{Nguyen2023ClimateLearnBM}
\bibinfo{author}{Nguyen, T.}, \bibinfo{author}{Jewik, J.}, \bibinfo{author}{Bansal, H.}, \bibinfo{author}{Sharma, P.} \& \bibinfo{author}{Grover, A.}
\newblock \bibinfo{journal}{\bibinfo{title}{Climatelearn: Benchmarking machine learning for weather and climate modeling}}.
\newblock {\emph{\JournalTitle{ArXiv}}} \textbf{\bibinfo{volume}{abs/2307.01909}} (\bibinfo{year}{2023}).

\bibitem{Kaltenborn2023ClimateSetAL}
\bibinfo{author}{Kaltenborn, J.} \emph{et~al.}
\newblock \bibinfo{journal}{\bibinfo{title}{Climateset: A large-scale climate model dataset for machine learning}}.
\newblock {\emph{\JournalTitle{ArXiv}}} \textbf{\bibinfo{volume}{abs/2311.03721}} (\bibinfo{year}{2023}).

\bibitem{Martini_Kalaitzis_Chantry_Watson-Parris_Bilinski_2021}
\bibinfo{author}{Schroeder~de Witt, C.} \emph{et~al.}
\newblock \bibinfo{journal}{\bibinfo{title}{Rainbench: Towards data-driven global precipitation forecasting from satellite imagery}}.
\newblock {\emph{\JournalTitle{Proceedings of the AAAI Conference on Artificial Intelligence}}} \textbf{\bibinfo{volume}{35}}, \bibinfo{pages}{14902--14910} (\bibinfo{year}{2021}).

\bibitem{Chen2020RainNetAL}
\bibinfo{author}{Chen, X.} \emph{et~al.}
\newblock \bibinfo{title}{Rainnet: A large-scale imagery dataset and benchmark for spatial precipitation downscaling}.
\newblock In \emph{\bibinfo{booktitle}{Neural Information Processing Systems}} (\bibinfo{year}{2020}).

\bibitem{cannon2015bias}
\bibinfo{author}{Cannon, A.~J.}, \bibinfo{author}{Sobie, S.~R.} \& \bibinfo{author}{Murdock, T.~Q.}
\newblock \bibinfo{journal}{\bibinfo{title}{Bias correction of gcm precipitation by quantile mapping: how well do methods preserve changes in quantiles and extremes?}}
\newblock {\emph{\JournalTitle{Journal of Climate}}} \textbf{\bibinfo{volume}{28}}, \bibinfo{pages}{6938--6959} (\bibinfo{year}{2015}).

\end{thebibliography}

\clearpage
\section*{Figure legends}

\textbf{Figure~\ref{fig:radar}.} \textbf{Map of ground-based radar stations.} The map shows the availability of precipitation data, with each blue dot representing a station. Coverage is relatively high in the Global North and comparatively low across the Global South. Image from the Tropical Globe radar database \cite{tropical_globe_radar_database}.

\medskip 

\noindent\textbf{Figure~\ref{fig:training_cfgs}.} \textbf{Illustration of training configurations.} The training configurations $A1,...,A4$ are composed of progressively larger subsets of the 12 selected training regions located in the Global North. The choice of regions is guided by availability of high-resolution observational data and inspired by existing works \cite{cooper2023analysiscgangenerativedeep, prasad2024evaluating}.

\medskip

\noindent\textbf{Figure~\ref{fig:rel_bilinear}.} \textbf{Heatmap of \% improvement relative to interpolation.} Change in CRPS (lower better) in $[\%]$ for each model relative to bilinear interpolation. $A_{1},...,A_{4}$ represent hierarchical training scenarios with progressively more high-resolution data, from training on a single region (A1) to using several training regions across the Global North (A4). Positive values show improvements over the interpolation baseline. Model performance generally improves when expanding the training areas from A1 to A4, but this trend depends strongly on the geographical region. The N/A value indicates an instance where the CRPS value is not reliable due to numerical instabilities during inference, likely driven by large distributional differences between training and target regions.

\medskip

\noindent\textbf{Figure~\ref{fig:rel_target}.} \textbf{Heatmap of \% performance drop between in and out-of-distribution training.} Change in CRPS (lower better) in $[\%]$ for each model relative to training the model directly on the target regions. $A_{1},...,A_{4}$  represent hierarchical training scenarios with progressively more high-resolution data, from training on a single region (A1) to using several training regions across the Global North (A4). Many regions show large negative values, indicating substantial drop in performance when evaluating on regions out-of-distribution, highlighting challenges in spatial generalization. 
The N/A value indicates an instance where the CRPS value is not reliable due to numerical instabilities during inference, likely driven by large distributional differences between training and target regions. 

\medskip

\noindent\textbf{Figure~\ref{fig:location_split}.} \textbf{Illustration of location splits and training configurations.} Patches $T_{1},...,T_{12}$ represent training regions and patches $E_{1},...,E_{6}$ correspond to evaluation areas that are used within $4$ sub-tasks, simulating different scenarios that correspond to varying levels of data availability.

\medskip

\noindent\textbf{Figure~\ref{fig:setup}.} \textbf{Graphical summary of RainShift setup.} The inputs of the downscaling model are a combination of ERA5 time series data and geographical features. The downscaling model is then able to generate probabilistic samples. For training, we sample from geographic areas $T_1,\ldots,T_{12}$ and years 2001--2020, and compare the generated samples with the ground truth target IMERG to compute the loss. For evaluation, we use areas $E_1,\ldots,E_6$, and years 2021--2022.

\medskip

\noindent\textbf{Figure~\ref{samples}.} \textbf{Qualitative comparison of downscaled precipitation fields.} This plot shows a sample, one time step from the evaluation set in the Cape Horn area in the first two columns and aggregated features in the last column. The random sample includes the matching input (ERA5) and target (IMERG) precipitation values in logarithmic scale and two samples each from the two generative approaches, GANs and DMs. The right column shows on top, the mean precipitation over the whole testing period (2021-2022) as well as the spatial distribution of CRPS scores for GAN and DM.

\medskip

\noindent\textbf{Figure~\ref{cdf}.} \textbf{Cumulative distribution functions (CDFs) of precipitation inputs for training and target regions.} For each target region, a mapping is constructed between the historic CDF of the training region and the historic precipitation inputs of the target region. This mapping is subsequently used to correct the future input data from the target region, aligning them more closely with the training data distribution. Since precipitation is highly right-skewed with many zero and near-zero values, which causes most values to fall within the first bin of the CDF, we show the logarithm of the CDF to better visualize the discrepancies in distributions.

\section*{Table legends}

\noindent\textbf{Table~\ref{tab:qm_results}.} \textbf{Impact of quantile-based domain alignment on model accuracy.} CRPS (lower values are better) of models trained on scenario A1 (Western North America) and evaluated on the designated target regions for precipitation in [mm/h] over test years $2021 - 2022$. Results show model accuracy with and without applying quantile mapping (QM) to align the input distributions between training and evaluation regions. All predictions are transformed back to the original target domain data range and re-normalized before computing the metrics. Applying quantile mapping equals or improves performance across most models and regions (except Cape Horn) demonstrating the potential of data alignment techniques to enhance spatial generalization under geographical distribution shifts.

\medskip 

\noindent\textbf{Table~\ref{tab:variables}.} \textbf{List of atmospheric variables used as predictors.}

\medskip

\noindent\textbf{Table~\ref{tab:stats}.} \textbf{Mean precipitation values.} Values are in mm/h averaged over each evaluation region using years 2021-2022.

\medskip

\noindent\textbf{Table~\ref{tab:crps_results}.} \textbf{Accuracy of models on spatial generalization tasks.}
Test CRPS (lower better) for precipitation in mm/h on the designated evaluation area and averaged over test years $2021 - 2022$. Shown is the mean, pixel-wise CRPS of 8 ensemble members. For deterministic models (ResNet and bilinear interpolation of ERA5 precipitation data), the MAE is shown. We report the mean precipitation amount in input and target data in mm/h. Best scores per subtask are in bold. The last three rows with scores in blue are not contestants in the benchmark but show what is possible when training directly on the target.

\section*{Author contributions statement}
P.H., D.R., C.L. and M.C. conceived the research. P.H., L.S., N.L., C.L., M.C. and D.R. designed the experiments. P.H. and F.P. implemented the code base and performed model training. P.H. and L.S. conducted the experiments and interpreted the results. P.H. and L.S. wrote the initial draft of the manuscript. P.H., A.H., and D.R. revised the manuscript.


\end{document}